\pdfoutput=1
\documentclass{article}

    \PassOptionsToPackage{numbers, compress}{natbib}

\usepackage[preprint]{neurips_2024}

\usepackage[utf8]{inputenc} 
\usepackage[T1]{fontenc}    
\usepackage{url}            
\usepackage{booktabs}       
\usepackage{amsfonts}       
\usepackage{nicefrac}       
\usepackage{microtype}      
\usepackage{xcolor}         
\usepackage{multirow}
\usepackage{multicol}
\usepackage{amssymb}
\usepackage[ruled,vlined]{algorithm2e}
\usepackage{colortbl}
\usepackage{longtable}
\usepackage{mdframed}
\usepackage{inconsolata}
\usepackage{color}
\usepackage{amsmath}

\definecolor{darkred}{rgb}{0.8, 0.0, 0.0}

\usepackage{subfigure}
\usepackage{array}
\usepackage{graphicx}
\usepackage{caption}
\usepackage{float}
\usepackage{adjustbox}
\usepackage{subcaption}
\usepackage{cancel}
\usepackage{enumitem}
\usepackage[most]{tcolorbox}

\definecolor{citecolor}{HTML}{0071BC}
\definecolor{linkcolor}{HTML}{ED1C24}
\usepackage[pagebackref=true,breaklinks=true,colorlinks,citecolor=citecolor, linkcolor=linkcolor]{hyperref}

\definecolor{fontgray}{RGB}{44, 62, 80}
\definecolor{myred}{RGB}{235, 47, 6} 
\definecolor{summertime}{RGB}{245, 205, 121}
\definecolor{darkgrass}{RGB}{0, 148, 50}
\definecolor{myblue}{RGB}{0, 168, 255}
\definecolor{mygray}{RGB}{158, 158, 158}
\definecolor{puffin}{RGB}{250, 152, 58}
\definecolor{lowpurple}{RGB}{210, 180, 222}
\definecolor{lowblue}{RGB}{102,178,255}
\definecolor{lowred}{RGB}{245, 183, 177}
\definecolor{deeppurple}{RGB}{142, 68, 173}
\definecolor{nephritis}{RGB}{39, 174, 96}

\definecolor{deepblue}{RGB}{41, 128, 185}
\definecolor{shymoment}{RGB}{162, 155, 254}
\definecolor{firstdate}{RGB}{250, 177, 160}
\definecolor{mintleaf}{RGB}{0, 184, 148}
\definecolor{alizarin}{RGB}{231, 76, 60}
\definecolor{soaring}{RGB}{149, 175, 192}
\definecolor{electronblue}{RGB}{9, 132, 227}
\definecolor{pinkgla}{RGB}{0, 184, 148}
\definecolor{coral}{RGB}{255, 127, 80}

\newcommand{\squishlist}{
\begin{list}{$\bullet$}
{   \setlength{\itemsep}{0pt}
   \setlength{\parsep}{3pt}
   \setlength{\topsep}{3pt}
   \setlength{\partopsep}{0pt}
   \setlength{\leftmargin}{1.5em}
   \setlength{\labelwidth}{1em}
   \setlength{\labelsep}{0.5em} } }
\newcounter{Lcount}
\newcommand{\squishlisttwo}{
\begin{list}{\arabic{Lcount}. }
  { \usecounter{Lcount}
 \setlength{\itemsep}{0pt}
 \setlength{\parsep}{0pt}
 \setlength{\topsep}{0pt}
 \setlength{\partopsep}{0pt}
 \setlength{\leftmargin}{2em}
 \setlength{\labelwidth}{1.5em}
 \setlength{\labelsep}{0.5em} } }
\newcommand{\squishend}{\end{list} }

\definecolor{lightpink}{rgb}{0.945, 0.816, 0.804}
\definecolor{lightgreen}{rgb}{0.851, 0.906, 0.839}
\definecolor{lightblue}{rgb}{0.8, 0.9, 1}
\definecolor{lightyellow}{rgb}{0.992, 0.949, 0.816}

\newtcolorbox[list inside=prompt,auto counter,number within=section]{prompt}[1][]{
    colbacktitle=black!60,
    coltitle=white,
    fontupper=\footnotesize,
    boxsep=5pt,
    enhanced,
    left=0pt,
    right=0pt,
    top=0pt,
    bottom=0pt,
    boxrule=1pt,
    breakable,
    #1
}

\begin{document}

\title{Citrus: Leveraging Expert Cognitive Pathways in a Medical Language Model for Advanced Medical Decision Support}
\author{Guoxin Wang$^{1}$\footnotemark[1], Minyu Gao$^{1}$, Shuai Yang$^{1}$, Ya Zhang$^{1}$, Lizhi He$^{1}$, Liang Huang$^{1}$, \\
\textbf{Hanlin Xiao}$^{1}$\footnotemark[2], \textbf{Yexuan Zhang}$^{1}$, \textbf{Wanyue Li}$^{1}$, \textbf{Lu Chen}$^{1}$, \textbf{Jintao Fei}$^{1}$, \textbf{Xin Li}$^{1}$ \\  
$^1$ Citrus Team, JD Health International Inc. \\
\url{https://github.com/jdh-algo/Citrus}}

\renewcommand{\thefootnote}{\fnsymbol{footnote}}
\footnotetext[1]{Project Lead}
\footnotetext[2]{Work done during the internship at Citrus Team.}

\maketitle

\begin{abstract}
Large language models (LLMs), particularly those with reasoning capabilities, have rapidly advanced in recent years, demonstrating significant potential across a wide range of applications. However, their deployment in healthcare, especially in disease reasoning tasks, is hindered by the challenge of acquiring expert-level cognitive data. In this paper, we introduce Citrus, a medical language model that bridges the gap between clinical expertise and AI reasoning by emulating the cognitive processes of medical experts. The model is trained on a large corpus of simulated expert disease reasoning data, synthesized using a novel approach that accurately captures the decision-making pathways of clinicians. This approach enables Citrus to better simulate the complex reasoning processes involved in diagnosing and treating medical conditions. To further address the lack of publicly available datasets for medical reasoning tasks, we release the last-stage training data, including a custom-built medical diagnostic dialogue dataset. This open-source contribution aims to support further research and development in the field. Evaluations using authoritative benchmarks such as MedQA, covering tasks in medical reasoning and language understanding, show that Citrus achieves superior performance compared to other models of similar size. These results highlight Citrus’s potential to significantly enhance medical decision support systems, providing a more accurate and efficient tool for clinical decision-making.
\end{abstract}

\begin{figure*}[ht!]
    \centering
    \includegraphics[width=0.95\textwidth]{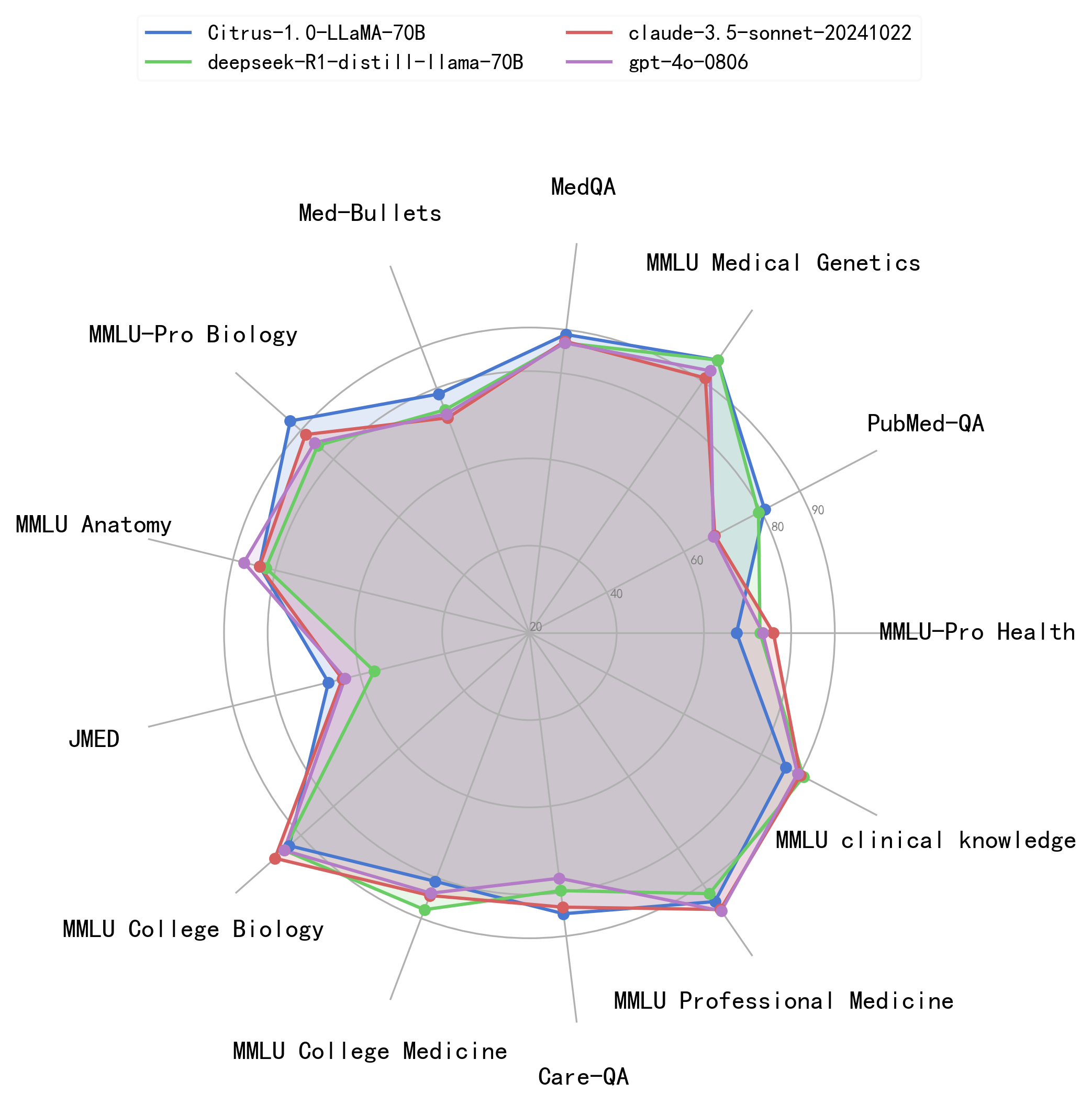}
    \caption{Citrus rank high in several authoritative medical benchmarks, comparing with two widely used LLMs, GPT-4o and Claude and a powerful 70B scale LLM, which is distilled from DeepSeek-R1.}
    \label{pic0-radar}
\end{figure*}

\section{Introduction}
Recent advancements in the reasoning capabilities of LLMs have become a focal point in research and are increasingly seen as a benchmark for assessing the intelligence level of these models\cite{xie2024preliminary, guo2025deepseek}. While the progress in reasoning capabilities has been rapid in domains like mathmetics and programming, the development in healthcare remains relatively limited\cite{hua2024large,qiao2022reasoning, ahn2024large, huang2022towards}. The open-ended nature of medical practice presents a more complex challenge for Language Models. Medical expertise is cultivated through real-world clinical practice, making it essential for medical reasoning models to learn from the diagnostic and treatment processes of human experts. As a result, emulating the reasoning pathways of medical professionals becomes a crucial step for developing effective medical reasoning models. 

Clinical practice, requiring highly sophisticated medical reasoning skills, encompasses patient consultation, diagnosis, differential diagnosis, and treatments\cite{symons2017differential,norman2005research}. Medical experts have systematically summarized the thought processes involved in clinical practice\cite{higgs2008clinical,adams2013clinical,schwartz2008clinical,schmidt1990cognitive}.  For medical language models to successfully assist in clinical decision-making, they must not only process vast amounts of medical data but also emulate the complex cognitive processes of expert medical professionals\cite{monteiro2013diagnostic}. This requires LLMs to understand not only the explicit medical knowledge but also the implicit reasoning steps that experts use when diagnosing and treating patients. Furthermore, as medical decisions often involve ambiguity, incomplete data, and uncertainty, models must be able to handle these complexities in a way that mirrors expert judgment. 

To achieve this, it is necessary for medical language models to be trained on data that closely mirrors the decision-making processes of clinicians. However, obtaining real-world expert-level clinical reasoning data is challenging, as it requires capturing the nuances of expert thought, which are often difficult to quantify. Moreover, existing datasets, although valuable, often fail to replicate the dynamic and often ambiguous nature of clinical practice. In response, new approaches to data synthesis and model training are required to bridge these gaps, enabling models to mimic expert reasoning while also adapting to the complexities and variability inherent in medical practice. 

In collaboration with human medical experts, we have designed a methodology for enhancing complex reasoning capabilities in large medical language models, specifically tailored for clinical scenarios. This training-free approach emulates the cognitive reasoning processes of medical professionals, resulting in significant improvements in complex reasoning tasks across multiple models, including Llama3.1\cite{dubey2024llama} and GPT-4\cite{achiam2023gpt}. Inspired by the results, we took further steps and carefully designed a multi-stage training method incorporating multiple phases of continuous pre-training (CPT), supervised fine-tuning (SFT) and reinforcement learning (RL). In this paper, we present Citrus, a medical language model that leverages expert cognitive pathways to simulate the reasoning processes of clinicians. By training Citrus on a large corpus of simulated medical reasoning data, we replicate the dynamic and iterative nature of clinical decision-making. This enables the model to engage in more accurate and effective reasoning, forming the foundation for future AI-driven medical decision support systems. Additionally, we successfully translated this methodology into a trainable approach, resulting in substantial performance improvements in several open-source base models across a variety of medical benchmarks. The advanced medical reasoning and decision-making support provided by Citrus have enabled it to outperform 70B parameter models in the medical domain. 

At the same time, we identified a key challenge in current medical language model evaluations: the structured nature of assessment questions often fails to capture the inherent ambiguity of patient symptoms in real-world clinical practice. By leveraging real-world consultations at JD Health’s internet hospital, we have created a clinical practice evaluation dataset  JDH MEDical Practice Dataset (JMED) that reflects real-world disease distribution, and can be regularly updated. 

The contributions of this paper are as follows:

\begin{enumerate}
\item We propose a training-free reasoning approach that emulates the cognitive processes of medical experts, enabling large language models to enhance their medical capabilities in clinical diagnosis and treatment.
\item In conjunction with the data construction method, we introduce a multi-stage post-training approach to further improve the model’s medical performance.
\item We have made the Citrus model and its training data publicly available as open-source resources to advance research in AI-driven medical decision-making.
\item We have developed and open-sourced a large-scale, updatable clinical practice evaluation dataset based on real-world data, accurately reflecting the distribution of patients in real-world settings.
\end{enumerate}

\section{Related Works}
\paragraph{Medical reasoning in clinical practice} In clinical practice, determining the most rational expert thinking process has always been a key research focus\cite{norman2005research,higgs2008clinical,adams2013clinical,schwartz2008clinical,schmidt1990cognitive}. The hypothetico-deductive method\cite{elstein1990medical,patel1986knowledge,higgs1992developing} is a reasoning process from general to specific, which determines diseases based on symptom combinations according to known medical theories. According to this method, some diagnostic hypotheses or conclusions has been raised firstly after collecting information from patients and will be waiting for testing. And these hypotheses, to some extent, guided the subsequent diagnosis and treatment. Pattern-recognition method\cite{barrows1987clinical,case1988evaluating} is a reasoning process from specific to general, which discovers patterns based on clinical observations and empirical summaries. Physicians quickly establish preliminary diagnoses through certain typical descriptions and specific combinations of symptoms that have been repeatedly validated in long-term clinical practice. The dual-processing theory (DPT)\cite{epstein1994integration,hammond2000human}, which integrates hypothesis testing methodologies with pattern recognition approaches, has gained widespread recognition and acceptance among medical experts. DPT includes system 1 and system 2\cite{evans2008dual}.  System 1 is a fast, intuitive, non-analytical process which is similar to pattern-recognition method, while System 2 is a slow, deliberate, analytical process related to hypothetico-deductive method\cite{higgs2008clinical,pelaccia2011analysis}. DPT posits that the reasoning pathway in clinical practice necessitates the concurrent integration of both intuitive and analytical processes\cite{evans2008dual,kahneman2011thinking,evans2013dual}.

\paragraph{Application of Large Language Models in Medical Reasoning} Researchers have realized the great potential of LLMs reasoning in medical problems solving\cite{zhou2023survey,karttunen2023large,hua2024large}. Recent advancements in LLMs for healthcare have been propelled by improved training methodologies, including CPT, SFT, and RL, which significantly enhance medical dialogue comprehension and question-answering capabilities\cite{zhou2023survey,lee2020biobert,alsentzer2019publicly,peng2023study,xiong2023doctorglm,li2023chatdoctor,han2023medalpaca,ye2023qilin,yang2024zhongjing,wu2024clinical,singhal2025toward,singhal2023large,MedicalGPT,wang2023huatuo,wu2024pmc,bao2023disc,zhang2023biomedgpt,chen2024huatuogpt,wang2024apollo,zheng2024efficiently,christophe2024med42}. Training-free techniques, such as advanced prompt engineering, have enabled general-purpose LLMs to perform specific medical tasks without retraining, as evidenced by studies like MedPrompt\cite{liu2023deid,singhal2023large,nori2023can,bayarri2024prompt,liu2024medcot,wu2024knowlab_aimed,lievin2024can,saab2024capabilities}. The implementation of multi-agent systems simulating experts from various medical departments has improved decision-making and overall medical performance by supporting complex tasks such as multi-step reasoning and treatment planning\cite{chen2024cod,tang2023medagents,schmidgall2024agentclinic,li2024agent}. Research has highlighted the potential of generating intermediate steps to enhance reasoning abilities, exemplified by OpenAI’s O1\cite{xie2024preliminary}. Additionally, R1 has demonstrated that training with large-scale synthetic data can yield exceptional reasoning models\cite{guo2025deepseek}. Inspired by these innovations, models such as Huatuo-O1\cite{chen2024huatuogpto1medicalcomplexreasoning}, O1-Journey\cite{qin2024o1,huang2025o1}, and Baichuan-M1\cite{wang2025baichuan} have been developed, utilizing inference-time scaling to produce extended reasoning outputs, thereby excelling in diagnostic tasks involving complex medical cases\cite{yuan2023scaling}. Huatuo-O1 focuses on advancing the complex reasoning capabilities of LLMs in healthcare by constructing verifiable medical problems and employing medical validators to ensure output accuracy. In contrast, O1-Journey emphasizes enhancing LLMs' ability to handle intricate medical tasks through reasoning augmentation. Baichuan-M1, developed from scratch and specifically optimized for medical applications, is designed to excel in both general domains such as mathematics and programming, as well as specialized medical fields including diagnostic support, medical research, and treatment recommendations. Building on these advancements, our objective is to effectively emulate doctors' cognitive processes in clinical practice to enhance the medical capabilities of large language models.

\paragraph{Evaluation of medical capabilities in large language models} Large language models have shown considerable promise in the medical field, and several benchmarks exist to evaluate their capabilities in this domain. Some studies compile medical license examination questions into medical competency assessment datasets, evaluating large language models’ medical capabilities in the same way medical students are tested\cite{huang2025o1}. Some works collect key questions from medical papers, requiring large language models to read medical paper abstracts to answer medical research questions, examining the models’ ability to comprehend medical literature\cite{BenAbacha:MEDINFO19}. Furthermore, to more accurately assess and differentiate the reasoning abilities of large models, the MMLU-Pro\cite{wang2024mmlu} dataset selects more challenging and reasoning-focused questions from MMLU\cite{hendrycks2020measuring} and expands the number of answer choices from four to ten. We aim to combine the advantages of these works to construct a clinical practice evaluation dataset that aligns with the distribution characteristics of real patients and can be regularly updated. 

\section{Training Data}
\subsection{Understanding Clinical Reasoning}
Recent studies have used the Chain-of-Thought (COT)\cite{wei2022chain} generation technique to enhance the reasoning capabilities of medical models. We argue that structuring the reasoning process to mirror the cognitive pathways of expert doctors in a structured COT approach is more effective in activating the model’s reasoning potential compared to unstructured, base-model-driven processes. Additionally, structured reasoning is easier for human experts to verify. Upon observing the reasoning processes of medical professionals, we identify two primary reasoning methods: the hypothetico-deductive method and the pattern-recognition method. The hypothetico-deductive method involves generating hypotheses based on available information, testing these hypotheses against further data, and revising them to form conclusions. This method emphasizes critical thinking and careful hypothesis testing, making it a robust approach in clinical practice, especially when facing complex, uncertain cases.In contrast, the pattern-recognition method relies on the recognition of patterns or symptoms that closely match previous cases or well-established medical knowledge. This method is often more intuitive and is useful when dealing with familiar or straightforward cases. It involves rapid decision-making based on experience rather than hypothesis testing. Experienced experts typically combine both methods in clinical decision-making to ensure efficient and accurate outcomes.

\begin{figure*}[ht!]
    \centering
    \includegraphics[width=0.95\textwidth]{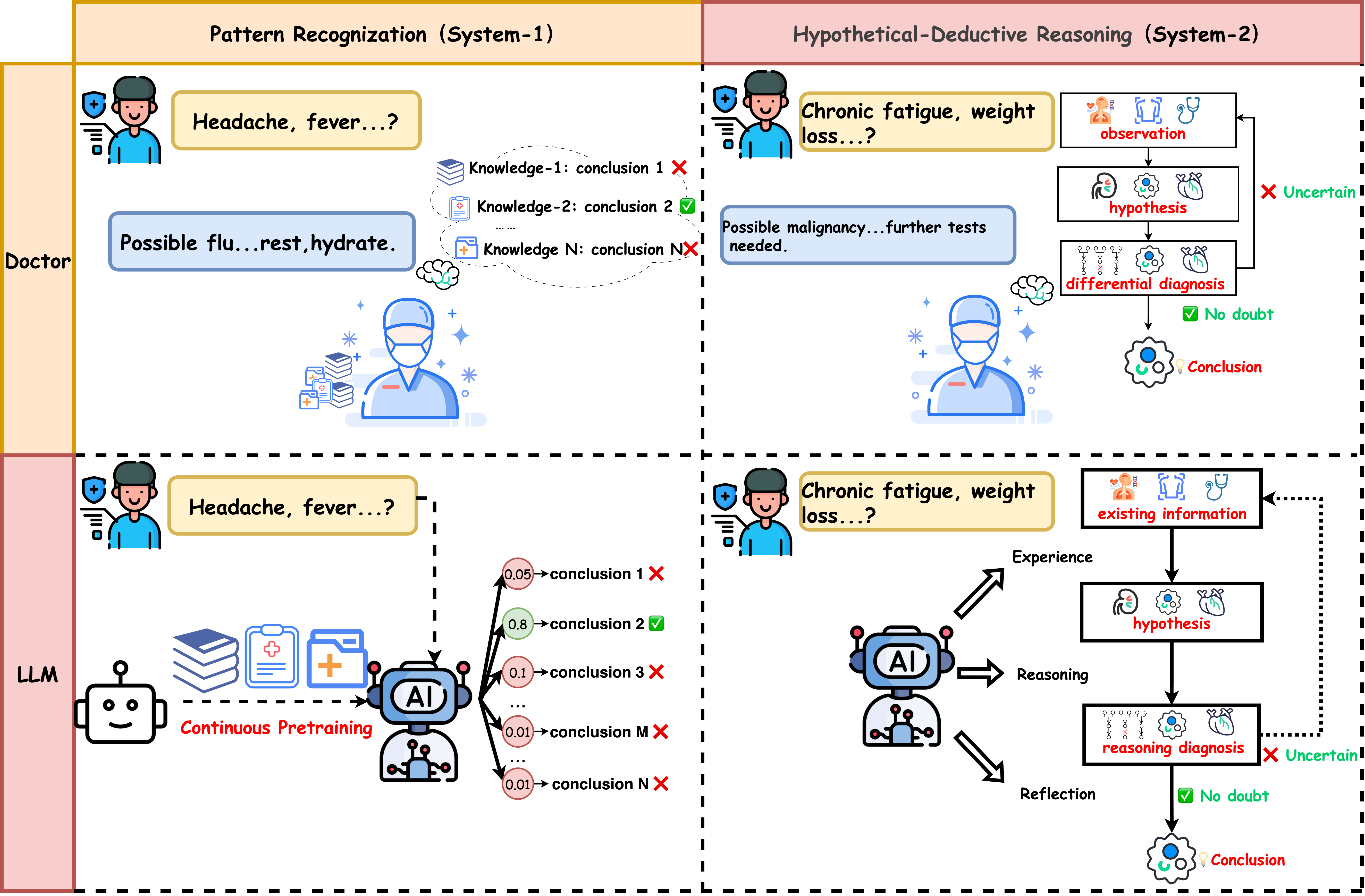}
    \caption{LLMs preforms similar cognitive pathways as medical experts. CPT enabled LLMs to learn medical knowledge and perform pattern recognition as doctors do, meanwhile LLMs are capable to handle hypothetical-deductive reasoning by executing several specific reasoning steps, which can be trained through SFT and RL procedure.}    
    \label{fig:Figure2}
\end{figure*}

Leveraging the cognitive pathways of medical experts, we propose a multi-stage data construction methodology that allows LLMs to integrate both reasoning methods, emulating expert reasoning patterns in medical decision-making\cite{wu2024clinical}, shown in Figure.\ref{fig:Figure2}. The main approaches are as follows:
\begin{enumerate}
    \item Pattern-recognition capabilities are typically developed through CPT. Through exposure to large-scale, high-quality medical datasets, LLMs can most intuitively learn the logical relationships and probability distributions among various medical entities.
    \item Hypothetical-deductive reasoning capability requires LLMs to manipulate medical knowledge sophisticatedly. To emulate this complex cognitive pattern, we synthesized extended COT data by simulating expert reasoning processes. Additionally, a two-stage curriculum learning strategy is implemented as a prerequisite to smooth the model’s learning trajectory.
\end{enumerate}

\subsection{CPT Data for Pattern Recognition}
The pattern-recognition method is typically embedded in the pre-training of large language models and is refined further during domain-specific training. Through the CPT process, a comprehensive medical domain dataset is used to enhance the pattern recognition capabilities of LLMs in addressing medical challenges. By collecting medical field data and preprocessing it, we have obtained a CPT dataset that enables LLMs to learn medical knowledge and perform pattern recognition.

\paragraph{Data Collection}The sources of CPT data include the following aspects:
\begin{itemize}
    \item Web Data
    \item Medical Textbooks 
    \item Medical guidelines and literature.
\end{itemize}

\paragraph{Data Process}
Data sourced from the web requires careful attention to data cleaning and labeling processes. Following the RedPajama\cite{weber2025redpajama} approach, we applied natural language processing techniques, such as entity recognition, relationship extraction, and text classification, for data cleaning and labeling. Additionally, we performed deduplication to ensure the quality of the dataset.

For handling PDFs with complex structures, we leveraged certain computer vision solutions. In contrast to web text, data from research papers and medical guidelines presents a different challenge. While this type of data often boasts high-quality content, extracting and transforming it from highly complex structured formats into a form suitable for the CPT training process is particularly challenging. We processed over one million PDF documents using methods such as MAPneo\cite{zhang2024map}.

Regarding medical textbooks, we applied data augmentation techniques to synthesize data. Data derived from medical textbooks inherently represents an optimal training corpus. However, due to limitations on the number of medical books available for collection, this dataset is smaller compared to web and academic materials. In order to ensure that these high-quality data points have a significant impact during training without being overshadowed by the other two data sources, we utilized techniques such as WizardLM’s self-evolution method to diversify and expand the knowledge from individual books into a variety of medical queries\cite{xu2023wizardlm}.

\paragraph{Scale and Distribution}
To mitigate the performance degradation caused by catastrophic forgetting, we cannot train the model solely on medical corpora. Therefore, we cleaned and selected approximately 200 billion tokens of CPT corpus from the following public datasets (CCI\cite{wang2024cci30hqlargescalechinesedataset}, PubMed\cite{pubmed}, SlimPajama\cite{cerebras2023slimpajama}, WuDao\cite{c6a3fe684227415a9db8e21bac4a15ab}, ChineseWebText\cite{chen2023chinesewebtext}, Math Pile\cite{wang2023generative}, Stack Code\cite{lozhkov2024starcoder}, etc.) and purchased medical book data. After categorizing and labeling the data, we found that medical data accounted for about 30\% of the total.

\subsection{Data Synthesis for Hypothetico-Deductive Reasoning}
The hypothetico-deductive method is typically characterized by the following steps in an expert's thinking process: collecting information, analyzing symptoms, generating diagnostic hypotheses, conducting differential diagnosis, and reaching conclusions. In this process, hypothesis generation and hypothesis testing are the core components of the reasoning process. To model this, we propose a comprehensive data series including general ability sft data with data course and medical ability sft data inspired from training-free dual expert data synthesis system. 

\subsubsection{General Medical Instruction Data}
To enhance the model's fundamental instruction-following capabilities and improve its ease of training, we design a two-stage data course that is not limited to medical problems. We refer to these as Stage-1 and Stage-2 SFT Data\cite{InfinityInstruct2024}. We recognize that it is an impractical task to directly train a base model, which has undergone medical knowledge CPT, to acquire medical reasoning abilities. LLMs struggle to effectively address complex medical reasoning problems when starting with no prior task-handling capabilities. To address this, we adopted a two-stage, general-purpose SFT data approach as part of our data curriculum.

In the first stage, we train the model with approximately 7 million basic instruction examples to improve its ability to follow simple instructions. In the second stage, we use 1.4 million higher-quality and more complex instructions, aiming to enhance the model’s multi-turn dialogue handling and complex instruction-following capabilities while preserving the abilities gained from the first stage. This process results in the stage-2 SFT model, which provides a solid foundation for more specialized task training in subsequent phases.

\subsubsection{Dual Expert Reasoning Method}
In this section, we present the Dual-Expert Reasoning Method. Through this approach, LLMs can emulate medical experts by employing hypothetico-deductive reasoning processes to address medical problems.

To emulate the Hypothetico-Deductive Process , we established a Reasoning Expert. When confronted with a problem, this role analyzes the available information, formulates new hypotheses, and conducts thorough reasoning. During the Training-free experiments, we observed that this process allows considerable flexibility. When the model does not engage in reflection, a significant amount of invalid reasoning processes are generated. This is unacceptable in terms of both reasoning accuracy and training efficiency. To address this, a multi-expert ensemble approach proves to be an effective solution. Thus, we designed a second expert, called the Reflection Expert. The Reflection Expert is tasked with evaluating the reasonableness of the reasoning process and discarding unreasonable or irrelevant steps. We then designed a cognitive flow loop to ensure the model generates a sufficient number of reasonable and accurate reasoning steps:
\begin{enumerate}
    \item The model lists the existing information as the starting point for reasoning.
    \item Based on the existing information, the model proposes possible diagnoses as the endpoints of the reasoning.
    \item Perform forward reasoning, attempting to establish the logical path from the starting point to the endpoint.
    \item Use the Reflection model to evaluate the validity of the reasoning. 
    \item Repeat steps 3-4: the model’s prior logical path and reasoning feedback should be visible to the model, which will attempt to establish more distinct logical paths. If all reasoning endpoints (diagnoses) have been fully discussed, and then rank the reasoning endpoints (diagnoses). Determine if a diagnosis can be made.
    \item If a diagnosis is made, output the result and conclude the reasoning.
    \item If a diagnosis cannot be made, return to step 1 and attempt to request external knowledge to gather more information.
\end{enumerate}

This method allows the model to emulate the structured, logical reasoning used by physicians in medical decision-making. By using dual-expert reasinong method, the model can generate multiple possible conclusions, evaluate the validity of each, and progressively refine its reasoning to arrive at the most plausible conclusion. Furthermore, when faced with incomplete or ambiguous information, the model can request external knowledge to assist in making a diagnosis, mimicking the diagnostic approach of medical professionals.

\subsubsection{Medical Reasoning Instruction Data}
In this section, we describe the construction of Stage-3 SFT Data using the Dual-Expert Reasoning Method. This dataset, called Citrus\_S3, designed to improve medical reasoning abilities in LLMs, will be open-sourced to promote further research and development in the field. To ensure accuracy and diversity, we propose several advanced data processing techniques, outlined below.

\paragraph{Reasoning Model with Ground Truth}
The key to generating reliable medical COT training data using this dual-expert method without additional supervised training is ensuring that the model generates a reasonable and accurate reasoning process. To achieve this, we modified the training-free method by providing the reflection model with the ground truth for the medical questions faced by the reasoning model. In this setup, the reflection model evaluates the reasonableness of the steps generated by the reasoning model and subtly guides it toward the correct answer, without directly providing the solution.This design results in a redefined step 4 in the dual-expert method.

\paragraph{Question Seeds}Another indispensable part to successfully execute this data generation procedure is to have extensive medical questions, which should be complicated enough to ignite reasoning process as well as equipped with ground truth that has been properly verified. 

\paragraph{Question Rewriting} 
The training question seeds in datasets like MedQA\cite{jin2021disease} are closed-form questions. We believe that providing answer options limits the model's reasoning capacity, restricting its ability to explore different reasoning paths. To improve generalization, we made the following adjustments:
\begin{itemize}
    \item We removed the options from the original closed-form questions and converted them into open-ended questions. This allows the model to focus on reasoning and conclusions without predefined answers. 
    \item We created a prompt for rewriting the open-ended questions, removing dependencies on options (e.g., "Which of the following statements is incorrect?").
\end{itemize}

\paragraph{Question Quality Control} 
Simple medical questions can be answered based on the model’s existing knowledge, but they do not require complex medical reasoning. To filter data useful for learning reasoning abilities, we used models such as GPT-4o-2024-0513, Qwen2.5-7B\cite{yang2024qwen2}, and Llamba-3.1-8B\cite{bick2025llamba} to answer closed-form questions. If these models answered correctly, the data was categorized as easy data, which does not require reasoning. Otherwise, it was categorized as hard data. During the SFT data sampling stage, we used all the hard data and a small portion of easy data to ensure the quality of the training dataset.

\paragraph{Data rewriting} Data rewriting is essential to transform multi-expert problem analysis into a first-person thought process. We use LLMs to accomplish this task with several strict constraints:
\begin{itemize}
    \item Keep thought scale.
    \item Use narrative words for transition words such as furthermore,therefore, then ,wait\dots
    \item Discard duplicated steps 
    \item Keep the ground truth 
\end{itemize} 

\subsubsection{Data Analysis}

\begin{table*}[ht!] \centering 
\adjustbox{max width=1.0\linewidth}{
\begin{tabular}{llllc} \toprule
\textbf{DataSet Name} & \textbf{Training Stage} & \textbf{Scale} & \textbf{Field} & \textbf{Construction Method} \\
\midrule

Web Data & CPT & 287B & General & Open-Source \\
Medical Textbooks & CPT & 4.6B tokens & modern medicine &  Collect \\
Medical guidelines and literature & CPT & 73B tokens & modern medicine & Collect  \\
Infinity-Instruct-7M &  SFT stage 1 & 7M lines & General\& Instruction Fellowing & Open-Source and in-house data\\
Infinity-Instruct-gen & SFT stage 2 & 1.4M lines & General\& Instruction Fellowing & Open-Source and in-house data \\
Citrus\_S3 & SFT stage 3 & 60K lines & Long COT on Medical Reasoning & Data Synthesis \\
Citrus\_xpo & RL & 50K pairs & Long COT on Medical Reasoning & Rejection Sampling\\

\bottomrule
\end{tabular}}
\caption{Training Data statistics. Our training data is incorporating with CPT data, SFT data and RL data. The table shows the scale, field and construction method of each dataset.}
\label{tab:datastats}
\end{table*}

This data synthesis approach, which is shown in Table.\ref{tab:datastats}, enables LLMs to generate medical COT data that aligns with medical logic, thereby enhancing their medical capabilities without the need for additional model training. Furthermore, by utilizing data synthesized using the hypothetico-deductive method for model training, the model can acquire medical reasoning capabilities similar to those of doctors.

\section{Model Training}
\begin{figure*}[ht!]
    \centering
    \includegraphics[width=0.95\textwidth]{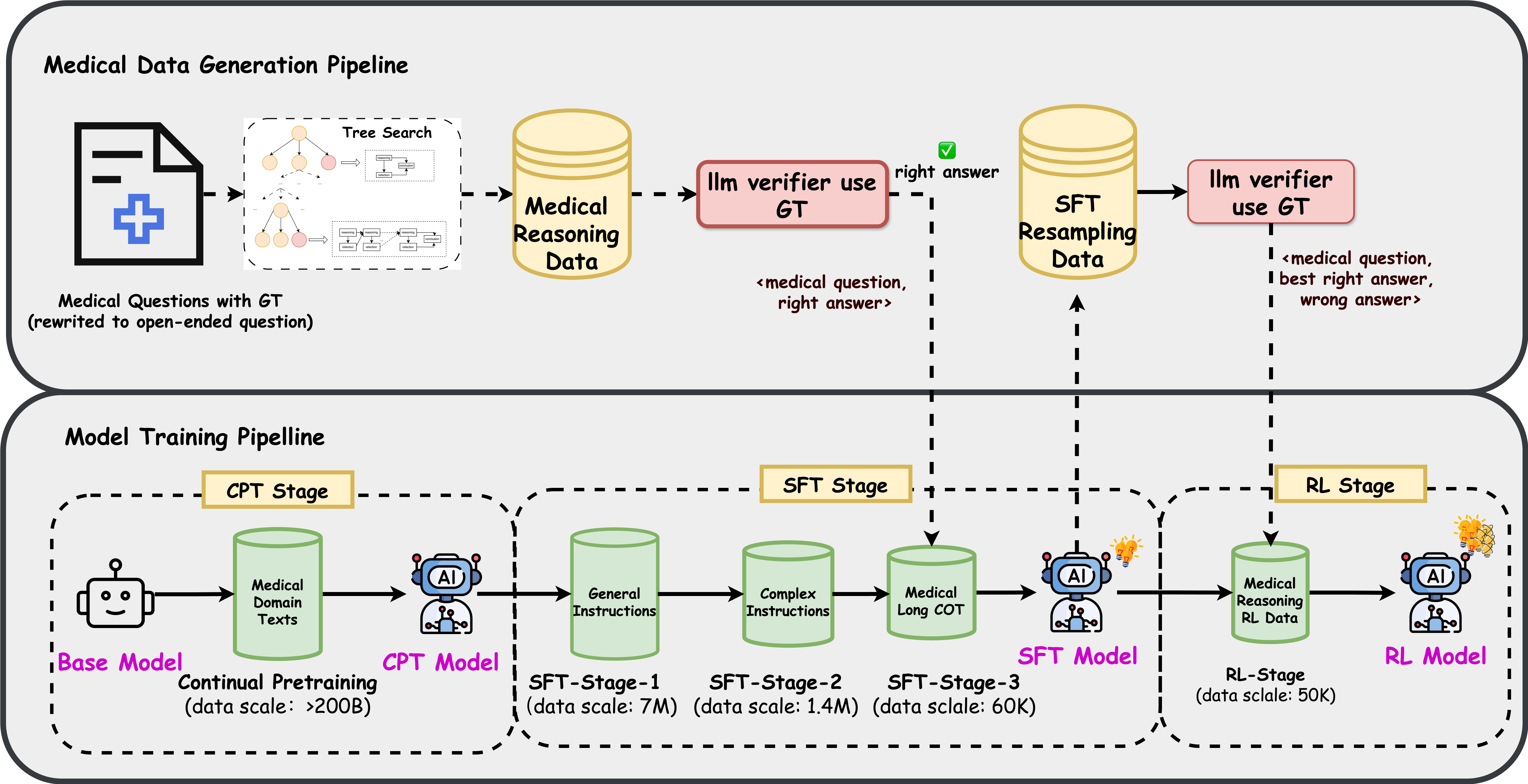}
    \caption{Overview of training stages and training data pipeline . The training process consists of three stages: CPT, SFT, and RL. We shows training purposes and dataset scale on each stage, also, we points out the data pipeline in corresponding stage.}
    \label{fig:Figure3}
\end{figure*}

In this section, we present a comprehensive training procedure that integrates multiple stages, including CPT, SFT, and RL, referred in Figure.\ref{fig:Figure3}. Through this multi-phase approach, we aim to transform a base model, initially lacking domain-specific medical knowledge and reasoning abilities, into a robust medical reasoning model capable of performing complex cognitive processes to effectively address and solve clinical problems. The training procedure leverages both general-purpose and medical-specific data, progressively refining the model’s ability to handle medical tasks and engage in sophisticated reasoning when confronted with real-world clinical scenarios.

\subsection{CPT Stage}
This phase focuses on the continuous pre-training of existing foundation models to enhance their comprehension of medical domain knowledge. A primary challenge lies in adapting the same dataset for different foundation models, which possess distinct training data ratios and quality control mechanisms.

In the continuous pre-training of large language models, the ratio of data from different sources is a critical topic. Here, we employ an AutoML approach to dynamically determine the proportion of each data source during training. Specifically, we frame the data ratio problem as a multi-armed bandit problem\cite{albalak2023efficient}. We hypothesize that the benefit of encountering previously seen content in continuous pre-training is relatively small, so the model should be encouraged to learn new knowledge. Therefore, we treat the training loss from each data source as a reward. Through this methodology, base models are exposed to training corpora with dynamically adjusted sampling ratios across different training phases, resulting in substantially improved convergence efficiency.

\subsection{SFT Stage}
\subsubsection{Medical Reasoning Ability SFT Training}
We propose a three-stage SFT training framework to enhance the model’s medical reasoning capabilities. As discussed in Section 3.3, the SFT datasets across these three stages are arranged in ascending order of difficulty. The underlying rationale is that the model should first master general knowledge application skills before proceeding to complex medical reasoning logic. In the following section, we will focus on elaborating the third stage of SFT training.

\begin{figure*}[ht!]
    \centering
    \includegraphics[width=0.95\textwidth]{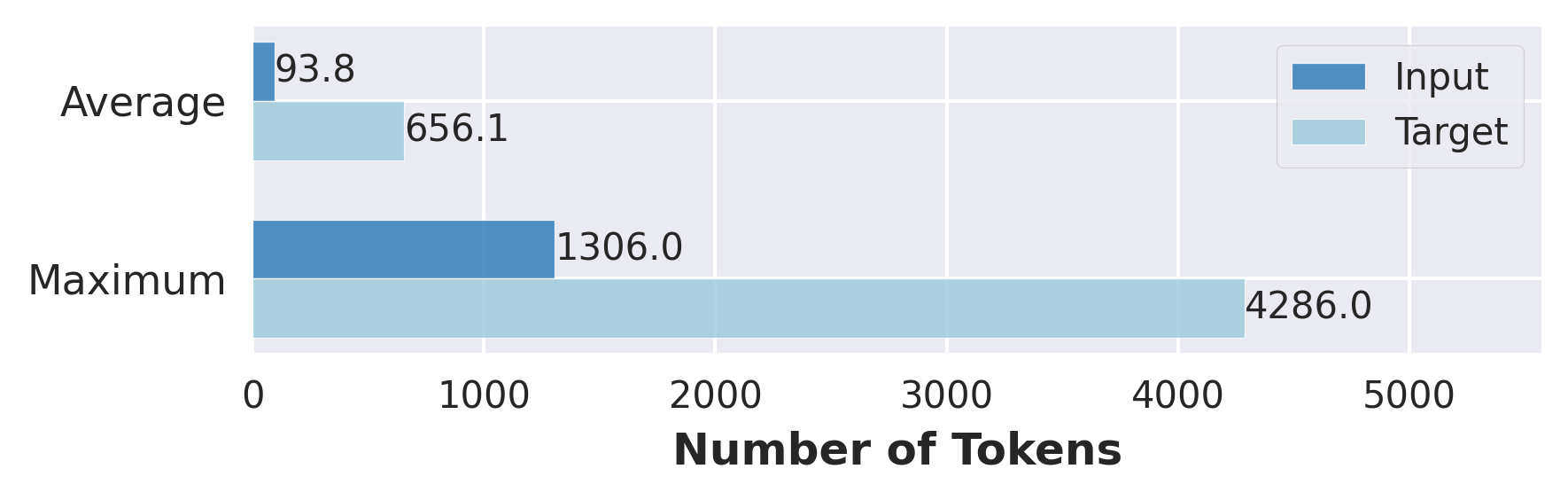}
    \caption {Token distribution statistics of stage 3 SFT training data. The data are designed and manufactured to simulate the long COT reasoning process of medical experts. We confirm a average length of 656 tokens and upper-bound of the length is around 4k.}
    \label{fig:Figure4}
\end{figure*}

The third phase of SFT training focuses on improving the model’s performance in the target task domain: medical reasoning. We used data obtained from the Training-Free approach and fine-tuned the Stage-2 SFT model in this phase. The main objective of this phase is to enhance the model’s ability to perform long COT in medical reasoning tasks. We used approximately 100,000 medical benchmark problems with GroundTruth gold-standard answers to generate reasoning data, which were used to train the model's medical reasoning capabilities. To maintain the model’s general-purpose capabilities during this process, we included reasoning data from other domains, such as logical and mathematical reasoning, in quantities comparable to the medical data. A toekn distribution statistics of stage 3 SFT training data is shown in Figure.\ref{fig:Figure4}.

\subsubsection{SFT Data Format} 
For all SFT data, we follow an unified template format: <sft-input, sft-target>. The sft-input consists of open-ended questions, and we expect the reasoning model to follow a structured thought process before providing answers. The resulting sft-target outputs follow the format: <think>thinking tokens</think><answer>answer tokens</answer>, where the reasoning process is contained within <think> and the final conclusion is given in <answer>.

\subsection{RL Stage}
After obtaining the Stage-3 SFT model, an effective reinforcement learning (RL) training phase is necessary. Compared to online methods, the RL techniques used in this phase, such as SIMPO\cite{meng2025simpo} and CPO\cite{abdel2024crested}, have distinct advantages. SimPO completely eliminates the dependency on reference models, by directly using the average log probability generated by the policy model as an implicit reward. This not only reduces computational and memory consumption but also simplifies the training process, avoiding the complexity brought by multi-stage optimization. By introducing length normalization, SimPO effectively prevents the model from being biased toward generating lengthy but low-quality responses due to the reward mechanism. However, SimPO is quite sensitive to the learning rate, so we introduced NLL loss, similar to the CPO method, to enhance training stability. These methods offer more stable and efficient learning compared to traditional online reinforcement learning methods. For the RL training, we used data that shares the same origin as that used in Pre-RL SFT, sampling and training on a dataset of approximately 50,000 instances.

\subsubsection{RL Data Sampling}
We use the best-performing checkpoint after the Stage-3 SFT to perform rejection sampling. The process is as follows: 

\paragraph{Repeat Sampling:} Open-ended questions (without answer options) are given to the model, which generates 20 responses at a high temperature (temperature = 1.2). 
\paragraph{Construct Preference Data:} To teach the model reasoning methods instead of just generating reasoning-like statements, we use rule-based rewards based on answer correctness, other than neural reward models. This ensures that rewards are accurately aligned with the correct reasoning steps. Specifically: 
\begin{itemize}
    \item Answer Mapping: For each response, we assess whether the open-ended answer corresponds to the correct option in the original closed-form question. Responses that align with the correct answer are classified as good responses, while others are considered bad responses. We only retain data that contains both good and bad responses. 
    \item Response Scoring: Each response is also scored using GPT-4o. The reasoning process and conclusion are assessed, and the highest-scoring good response is selected as the chosen response. For bad responses, the one with the lowest score is retained if multiple bad responses exist with the same incorrect option.
\end{itemize}

\subsubsection{RL Data Format} 
For the RL stage, the data format is <RL-input, chosen, rejected>. The RL-input format matches the sft-input format, and chosen and rejected follow the sft-target format.

\section{JDH Medical Practice Dataset: Construction and Validation of a Real-World Clinical Dialogue Benchmark}

Evaluating medical models is inherently challenging, especially when aligning them with real-world clinical settings. Effective evaluations should ensure that these models can be applied successfully in clinical practice.We systematically analyzed several widely-used medical QA datasets (e.g., MedQA\cite{jin2021disease}, PubMedQA\cite{jin2019pubmedqa}, MedMCQA\cite{pal2022medmcqa}, MedBullets\cite{chen2024benchmarking}, MMLU\cite{hendrycks2020measuring}, MMLU-Pro\cite{wang2024mmlu}, and CARE-QA\cite{ariasduart2025automaticevaluationhealthcarellms}), as shown in Table \ref{tab:JMED}. This analysis revealed three distinctive characteristics: (1) Most datasets are exclusively sourced from medical journal literature or professional medical examinations, with none incorporating real-world hospital data; (2) Question formats primarily consist of multiple-choice questions (MCQs) with 4-5 options, except for MMLU-Pro, which uses a 10-option format. These questions feature clear conditions and fixed options, failing to capture the ambiguity and limited diagnostic information encountered in real clinical settings; (3) With the exception of CareQA, the remaining datasets lack continuous updates after their initial release.

To address this, we developed the JMED, a novel dataset based on real-world medical data distributions. Unlike existing datasets, JMED closely mimics authentic clinical data while facilitating effective model training. Although based on real consultation data, it is not directly sourced from actual medical data, allowing us to incorporate key elements necessary for model training. We ensured compliance with ethical and legal standards throughout the data collection process, safeguarding privacy and meeting ethical guidelines. Due to the open-ended nature of medical consultations, where deﬁnitive answers are often elusive, the evaluation process is more challenging. To address this, each question includes 21 response options, with a "None of the above" choice. This design signiﬁcantly increases the complexity and diﬃculty of distinguishing the correct answers, thereby providing a more rigorous assessment framework.

Compared to existing medical QA datasets, JMED has three principal advantages: First, it more accurately reflects the ambiguity in patient symptom descriptions and the dynamic nature of clinical diagnosis in real-world scenarios. Second, the expanded response options require enhanced reasoning capabilities to identify the correct answers among numerous distractors. Additionally, leveraging the vast amount of consultation data from JDH Internet Hospital, we can continuously generate data that aligns with the distribution characteristics of real patients.

\begin{table*}[ht!] \small \centering
\adjustbox{max width=1.0\textwidth}{
\begin{tabular}{lp{2cm}p{2.5cm}p{3cm}p{2.5cm}p{3.5cm}} 
\toprule  
& \textbf{Data Source} & \textbf{Answer Format} & \textbf{Test Dataset Size} & \textbf{Released Time} & \textbf{Latest Update Time} \\
\midrule
MedQA & Examination & 4-option MCQs & 1,273 & 2022 & 2022 \\ 
PubMedQA & Literature & 3-option MCQs & 1,000 & 2019& 2019 \\
MedMCQA & Examination & 4-option MCQs & 4,183 & 2022 & 2022 \\
MedBullets & Examination & 5-option MCQs & 308 & 2024 & 2024 \\
MMLU & Examination & 4-option MCQs & 1,871 & 2021 & 2021 \\
MMLU-Pro & Examination & 10-option MCQs & 818 & 2024 & 2024 \\
CareQA & Examination & 4-option MCQs & 5,410 & 2020 & 2024 \\
\textbf{JMED} & Hospital & 21-option MCQs & 1,000 & 2025 & updatable \\
\bottomrule  
\end{tabular}}
\caption{Comparison of our dataset JMED with existing medical QA datasets. JMED outperforms the most amount of options in MCQs and is the only one based on real-world hospital data. These two factors make JMED more challenging and realistic.}
\label{tab:JMED}
\end{table*}

\subsection{Data Collection and Construction Pipeline}
\subsubsection{Raw Data Processing}
The dataset originates from anonymized doctor-patient dialogues at JD Health Internet Hospital, filtered to retain consultations adhering to standardized diagnostic workflows. The initial release contains 1,000 high-quality clinical records spanning all age groups (0-90 years) and multiple specialties.
\begin{itemize}
    \item Privacy Protection: Automated de-identification of sensitive information (names, institutions, locations) via regular expression matching.
    \item Data Balancing: Ensured statistical representativeness across age, gender, and medical specialties based on platform-wide consultation patterns.
    \item Deduplication: Applied semantic similarity algorithms to eliminate redundant chief complaints.
\end{itemize}

\begin{figure*}[ht!]
  \centering
  \includegraphics[width=0.95\textwidth]{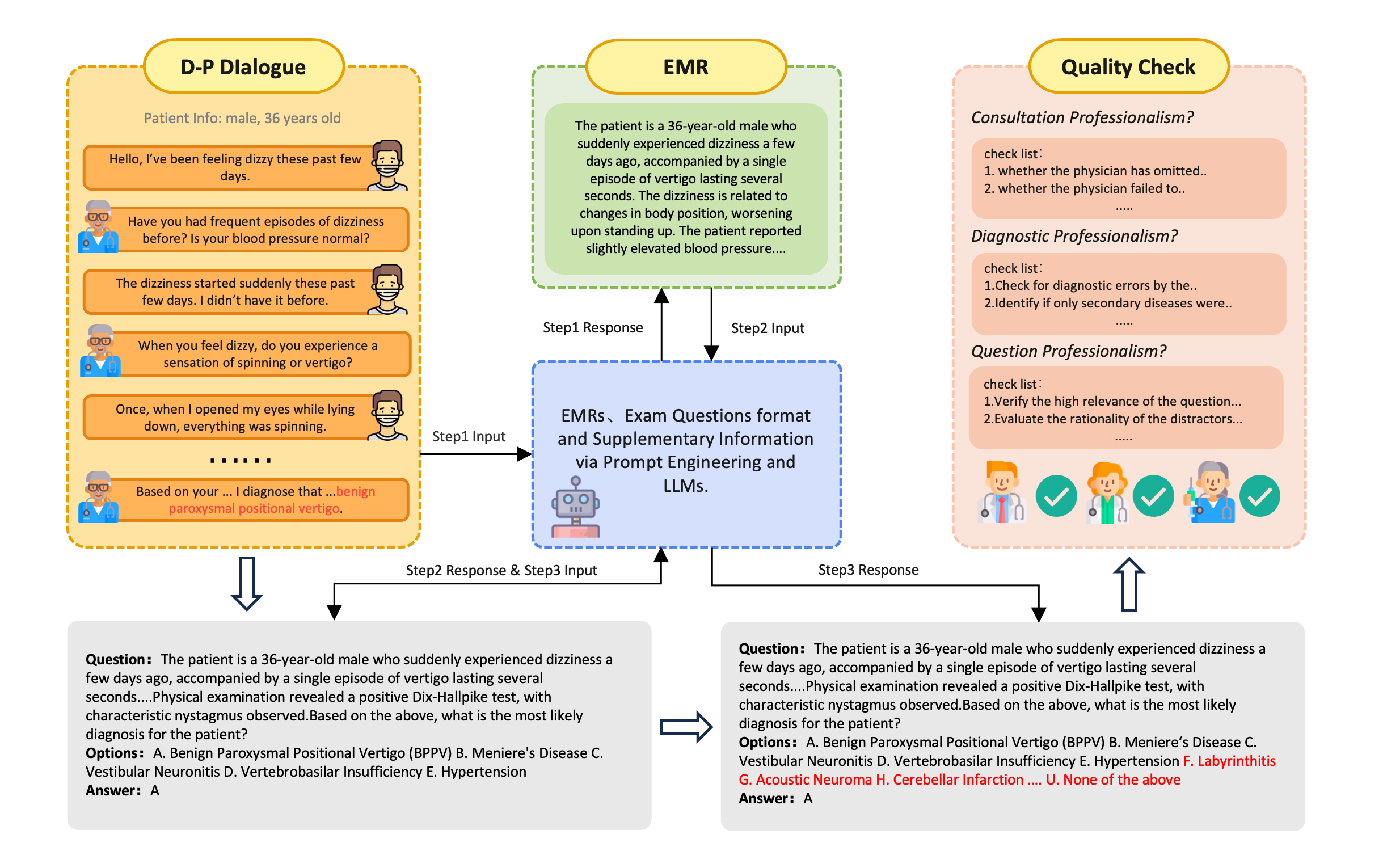}
  \caption{The construction framework of our dataset, JMED, is illustrated with arrows indicating the type of data used as input for the LLMs and the corresponding response obtained at each step. We begin by consolidating the dialogue data into EMRs, then transform it into the format of medical examination questions, and finally, through option expansion and quality checks, we obtain our dataset.}
  \label{fig:Figure5}
\end{figure*}

\subsubsection{Structured Transformation}
We constructed a set of multiple-choice questions (MCQs) based on the preprocessed data, as illustrated in Figure \ref{fig:Figure5}. 
\begin{itemize}
    \item Electronic Medical Record (EMR) Generation: Extracted key clinical elements using prompt engineering to create structured EMRs.
    \item Question Formulation: Employed the DeepSeek-r1 model to parse EMRs and generate clinically coherent questions aligned with diagnostic reasoning pathways.
    \item Option Expansion: Generated 21 mutually exclusive diagnostic options (including standardized ICD-10 terms and plausible differential diagnoses) using LLMs, ensuring compliance with the International Classification of Diseases, 10th Revision (ICD-10).
\end{itemize}

\subsection{Quality Assurance Framework}
Considering the seriousness and precision required in the medical field, a three-tier quality control system was established. This primary review process involves collaboration with physicians from 15 departments, with each department having two attending or associate attending-level doctors review the questions. Secondary validation is distributed to associate experienced physicians to conduct a re-evaluation, leveraging their expertise to ensure data quality and accuracy, and final audit is processed by chief physicians. All manual review processes must adhere to the criteria as describe in appendix \ref{Quality-Check:qc}.

Based on the aforementioned criteria, we have constructed a set of 1000 multiple-choice questions derived, encompassing multiple departments and age groups. Each data entry includes a unique ID, department, question, options, and the correct answer. The options adhere to the ICD-10 standard for disease nomenclature and have been reviewed and validated by professional physicians to ensure the appropriateness of the questions, options, and correct answers.

\subsection{Dataset Characteristics}
\begin{figure*}[ht!]
    \centering
    \includegraphics[width=0.95\textwidth]{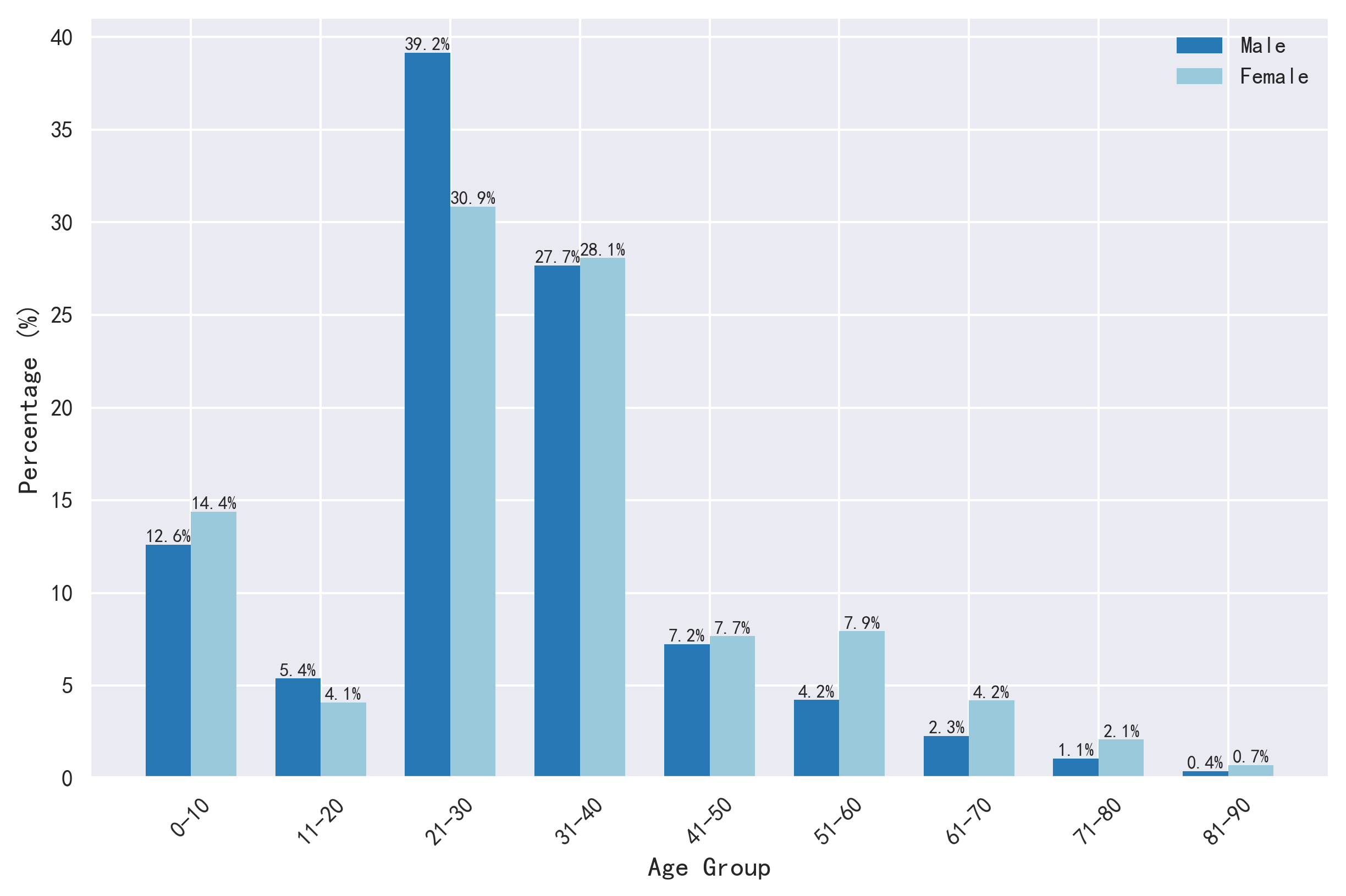}
    \caption{Age distribution statistics of males and females in our dataset. We got the main age distribution of the dataset is 21-40 years old with 83.38\% in portion. Meanwhile, we observe an unbalanced distribution in gender, that male patients are 58\%.}
    \label{fig:Figure6}
\end{figure*}
As shown in Figure \ref{fig:Figure6}, the validated subset comprises 1000 clinically rigorous multiple-choice questions with the following demographic distributions:
\begin{itemize}
    \item Age Coverage: Full spectrum (0-90 years), with 83.37\% of cases from the 21-40 age group after outlier adjustment.
    \item Gender Ratio: 58\% male vs. 42\% female.
    \item Specialty Distribution: Covers 15 clinical disciplines.
\end{itemize}

This distribution aligns with real-world usage patterns of online consultation platforms: adults constitute the primary user base, while consultations for pediatric and elderly patients are typically initiated by caregivers. Statistical analysis confirms that internet-based healthcare platforms have emerged as the preferred channel for adult populations seeking medical services.

\section{Experiments}
\subsection{Experimental Setup}
\paragraph{Implementation Details} We tested our methodology on Qwen2.5-72B\cite{yang2024qwen2}, LLama3.1-70B\cite{dubey2024llama} as our base models due to their foundational capabilities. The knowledge capacity of such large-scale models is a prerequisite for stimulating medical reasoning abilities. We use a two-stage SFT to enhance the general capabilities of the model, and in the third stage, we use a small amount of reasoning data to improve the model's medical reasoning ability. In the final stage, we performed reject sampling and alignment on the model to further enhance its reasoning capabilities. 
We use DeepSpeed ZeRO-3 and Accelerate to train the LLM, with AdamW as the optimizer. The $\beta1$ and $\beta2$ of AdamW are 0.9 and 0.95, respectively. We apply a weight decay of 1e-4 and clip the gradient norm to 1.0.
\paragraph{Training Hyperparameters}
The hyperparameter settings for our model training are shown in the table below. The training parameters for Qwen and LLaMA differ only in the learning rate during the alignment phase. Details are shown in Table \ref{tab:Table3}.
\begin{table*}[ht!] \small \centering
\adjustbox{max width=1.0\textwidth}{
\begin{tabular}{lp{3cm}p{2.5cm}p{2cm}p{2cm}p{4.5cm}} 
\toprule  
\textbf{model name} & \textbf{train phase} & \textbf{learning rate} & \textbf{Batch size} & \textbf{Epochs} & \textbf{Other hyperparameters} \\
\midrule
Citrus-Qwen-72B & stage1 & 5e-6 & 512 & 2 & - \\ 
Citrus-Qwen-72B stage1 & stage2 & 5e-6 & 512 & 1 & - \\
Citrus-Qwen-72B stage2 & stage3 & 5e-6 & 512 & 2 & - \\
Citrus-Qwen-72B stage3 & cpo-simpo & 1e-6 & 16 & 3 & $\alpha=0.05,\beta=10,\gamma=5.4$ \\
Citrus-Llama-70B & stage1 & 5e-6 & 512 & 2 & - \\
Citrus-Llama-70B stage1 & stage2 & 5e-6 & 512 & 1 & - \\
Citrus-Llama-70B stage2 & stage3 & 5e-6 & 512 & 2 & - \\
Citrus-Llama-70B stage3 & cpo-simpo & 3e-7 & 16 & 3 & $\alpha=0.05,\beta=10,\gamma=5.4$ \\
\bottomrule  
\end{tabular}}
\caption{Model Training Hyperparameter Settings. We show the training hyperparameters for both Qwen2.5-72B and Llama3.1-70B in different stages. We find only a slight difference in the learning rate during the alignment phase since Llama is harder to constrain with such a high learning rate which Qwen is working at.}
\label{tab:Table3}
\end{table*}

\subsection{Benchmarks} We utilized the \textbf{MedQA}\cite{jin2021disease}, \textbf{PubMedQA}\cite{jin2019pubmedqa}, \textbf{MedMCQA}\cite{pal2022medmcqa}, \textbf{MedBullets}\cite{chen2024benchmarking}, \textbf{MMLU}\cite{hendrycks2020measuring}, \textbf{MMLU-Pro}\cite{wang2024mmlu}, and \textbf{CARE-QA}\cite{ariasduart2025automaticevaluationhealthcarellms} as benchmarks, with \textbf{JMED} serving as a medical reasoning evaluation dataset specifically developed by us.

\textbf{MedQA} dataset is derived from multiple-choice questions of the United States Medical Licensing Examination (USMLE), covering English, Simplified Chinese, and Traditional Chinese. It is designed to evaluate a model's understanding and reasoning ability in medical knowledge.

\textbf{PubMedQA} is a biomedical question-answering dataset collected from PubMed abstracts, containing 1,000 expert-annotated, 61,200 unlabeled, and 211,300 artificially generated QA instances to evaluate a model's understanding and reasoning ability in biomedical research texts.

\textbf{MedMCQA} is sourced from multiple-choice questions in the AIIMS and NEET PG entrance exams. The dataset comprises over 194,000 multiple-choice questions, covering 2,400 healthcare topics and 21 medical subjects. Its purpose is to evaluate and improve models for generating answers to multiple-choice questions in the medical field.

\textbf{MedBullets} is a free learning and collaboration community that offers a large collection of USMLE-style questions and study resources. The dataset includes over 1,000 free USMLE Step 1-style questions, along with extensive study materials. The question type is primarily USMLE Step 1-style multiple-choice questions. Its purpose is to provide a learning and collaboration platform for medical students.

\textbf{MMLU} is a large-scale multitask language understanding benchmark dataset designed to evaluate the knowledge and reasoning abilities of large language models across multiple subjects.

\textbf{MMLU-Pro} is an improved and upgraded version of MMLU, designed to provide more challenging and difficult test questions.

\textbf{CareQA} is a healthcare QA dataset. The dataset originates from official sources of the Spanish Specialized Healthcare Training (FSE) examinations, including the biology, chemistry, medicine, nursing, pharmacology, and psychology tests from 2020 to 2024.

\textbf{JMED} dataset comes from JD Health's online internet hospital and is designed to simulate real clinical data.
\subsection{Main Results} 
We evaluated multiple open-source and closed source LLMs on medical tasks, as shown in Table below.

\begin{table*}[ht!]
\centering 
\adjustbox{max width=1\textwidth}{
\begin{tabular}{l>{\raggedright\arraybackslash}p{1.5cm}p{1.5cm}p{1.5cm}p{1.5cm}p{1.5cm}p{1.5cm}p{1.5cm}} \toprule
& \textbf{MedQA} & \textbf{PubMed-QA} & \textbf{Care QA} & \textbf{JMED} & \textbf{Med-Bullets} & \textbf{MMLU-Pro Health} & \textbf{MMLU-Pro Biology}\\
\midrule\midrule
\multicolumn{5}{c}{\bf \it \textbf{LLMs around 70B}} \\
deepseek-R1-distill-llama-70B & {0.8696} & 0.793 & \underline{0.7952} & {0.571} & 0.7468 & \textbf{0.7286} & \textbf{0.848}\\
Llama3.1-70B-instruct & 0.7722 & 0.793 & 0.5333& 0.559 & 0.6429 & 0.6467 & 0.7978 \\
huatuoGPT-o1-70B & 0.835 & \textbf{0.812} & 0.7095 & \text{-} & 0.763 & \underline{0.7164} & \underline{0.8382} \\
qwen2.5-72B-instruct & 0.7455 & 0.756 & - & \underline{0.667} & - & 0.665 & 0.834\\
O1-Journey Learning-llama-70B & 0.8648 & \text{-} & \text{-} & \text{-} & \underline{0.7727} & - & \text{-}\\
\rowcolor{orange!12} \textbf{Citrus1.0-llama-70B} & \textbf{0.8892} & \underline{0.809} & \textbf{0.8486} & \textbf{0.684} & \textbf{0.7857} & {0.6748} & {0.8326} \\
\midrule
\multicolumn{5}{c}{\bf \it \textbf{LLMs beyond 70B}} \\
claude-3.5-sonnet-20241022 & 0.8735 & 0.68 & 0.8333 & \text{0.669} & 0.7273 & \underline{0.7592} & \underline{0.8856} \\
gpt-4o-0513 & 0.8743 & 0.697 & 0.8095 & \text{0.668} & 0.7435 & \text{0.7323} & 0.8577 \\
gpt-4o-0806 & 0.8696 & 0.676 & 0.7667 & \text{0.644} & 0.737 & \text{0.7347} & 0.8577 \\
deepseek-v3 & 0.7824 & \underline{0.732} & 0.7667 & \text{0.646} & 0.6558 & \text{0.6993} & 0.8173 \\
deepseek-R1 & \underline{0.9097} & \textbf{0.767} & \textbf{0.9123} & \textbf{0.751} & \underline{0.8149} & {0.7518} & {0.8577} \\
gpt-o1-mini & 0.8955 & 0.706 & 0.6571 & \text{0.629} & 0.8084 & 0.7213 & 0.855 \\
gpt-o1-preview & \textbf{0.9513} & 0.725 & \underline{0.8714}& \underline{0.716} & \textbf{0.8896} & \textbf{0.7714} & \textbf{0.894} \\
\bottomrule
\end{tabular}}

\centering
\adjustbox{max width=1\textwidth}{
\begin{tabular}{l>{\raggedright\arraybackslash}p{1.5cm}p{1.5cm}p{1.5cm}p{1.5cm}p{1.5cm}p{1.5cm}p{1.5cm}} \toprule
& \textbf{MMLU Anatomy} & \textbf{MMLU clinical knowledge} & \textbf{MMLU College Biology} & \textbf{MMLU College Medicine} & \textbf{MMLU Medical Genetics} & \textbf{MMLU Professional Medicine} \\
\midrule\midrule
\multicolumn{5}{c}{\bf \it \textbf{LLMs around 70B}} \\ 
deepseek-R1-distill-llama-70B & 0.8222 & \textbf{0.9094} & \underline{0.9514} & \textbf{0.8786} & \textbf{0.96} & 0.9265 \\
Llama3.1-70B-instruct & 0.8148 & 0.8566 & 0.9306 & 0.7861 & \underline{0.95} & 0.9118 \\
huatuoGPT-o1-70B & \underline{0.837} & \underline{0.883} & \textbf{0.9583} & 0.8092 & \textbf{0.96} & \textbf{0.9632} \\
qwen2.5-72B-instruct & \textbf{0.8519} & 0.8528 & 0.9306 & \underline{0.8208} & 0.89 & 0.9044\\
O1-Journey Learning-llama-70B & \text{-} & \text{-} & \text{-} & \text{-} & \text{-} & \text{-}\\
\rowcolor{orange!12} \textbf{Citrus1.0-llama-70B} & \underline{0.837}  & {0.8642}  & {0.9357}  & {0.8092}  & \textbf{0.96}  & \underline{0.9485} \\
\midrule
\multicolumn{5}{c}{\bf \it \textbf{LLMs beyond 70B}} \\ 
claude-3.5-sonnet-20241022 & 0.837 & 0.9019 & \textbf{0.9792} & 0.8439 & 0.91 & \underline{0.9706}\\
gpt-4o-0513 & 0.8889 & 0.883 & \underline{0.9653} & 0.8555 & 0.96 & 0.9559 \\
gpt-4o-0806 & 0.8741 & 0.8943 & 0.9514 & 0.8382 & 0.93 & \textbf{0.9743}\\
deepseek-v3 & 0.837 & 0.8868 & 0.9514 & 0.8092 & 0.9 & 0.9301 \\
deepseek-R1 & \textbf{0.9259} & \textbf{0.9283} & \textbf{0.9792} & \textbf{0.8844} & \textbf{0.98} & {0.9596} \\
gpt-o1-mini & 0.8074 & 0.8604 & 0.9444 & 0.8439 & 0.96 & 0.9596 \\
gpt-o1-preview & \underline{0.9185} & \underline{0.9094} & 0.9514 & \underline{0.8728} & \underline{0.97} & \underline{0.9706} \\
\bottomrule  
\end{tabular}}
\caption{Main Results on Medical Benchmarks. LLMs are seperated into 70B scale group and beyond 70B group. Citrus leads most benchmarks among 70B LLMs, moreover, Citrus also surpasses several LLMs beyond 70B on medical benchmarks. \textbf{bold} highlights the best scores, and \underline{underlines} indicate the second-best.}
\label{tab:Table4}
\end{table*}

\begin{figure}[htbp]
    \centering
    \begin{minipage}{0.45\textwidth}
        \centering
        \includegraphics[width=\textwidth]{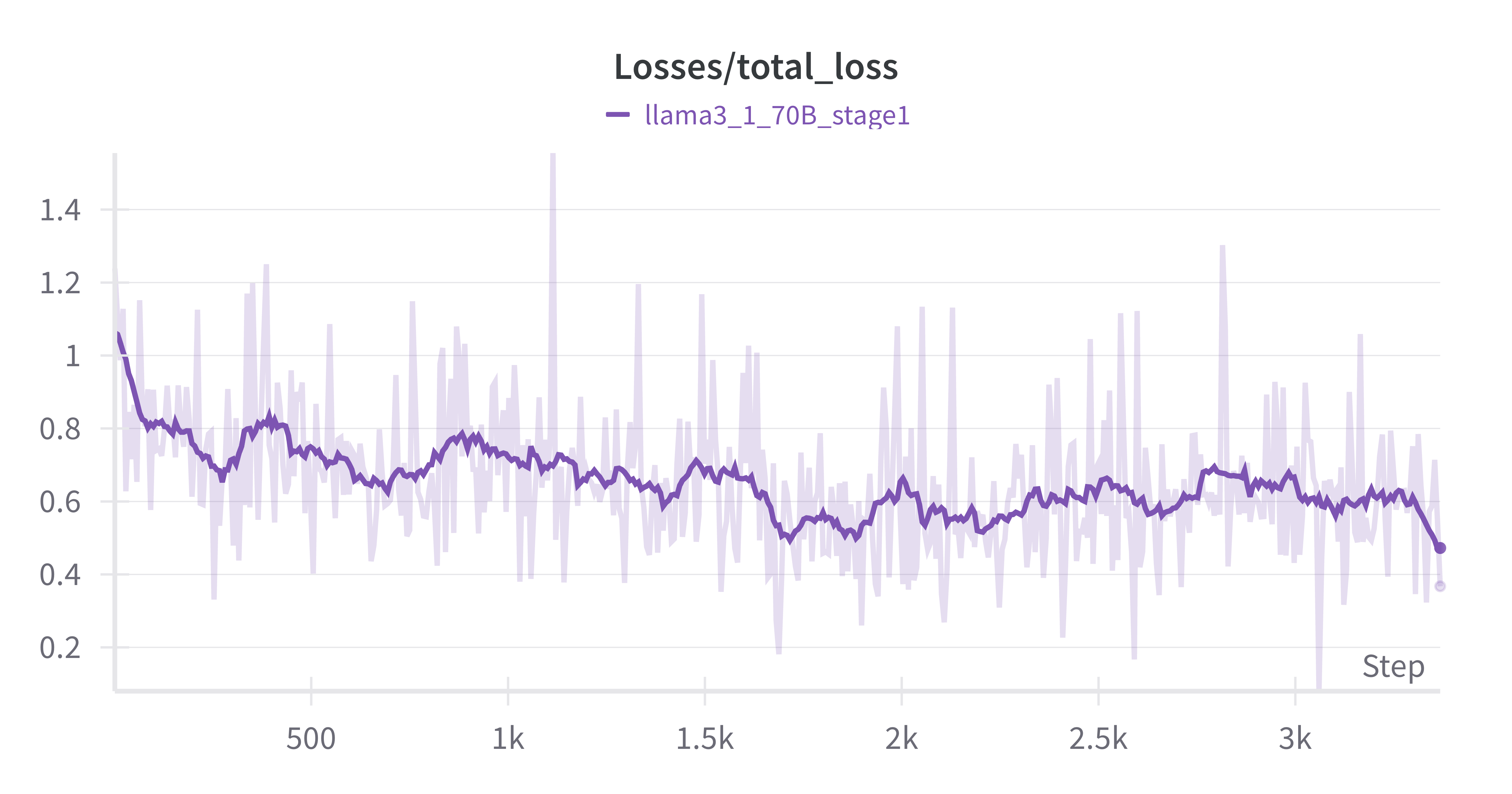}
        \caption*{(a) SFT Stage1 Loss}
    \end{minipage}
    \hfill
    \begin{minipage}{0.45\textwidth}
        \centering
        \includegraphics[width=\textwidth]{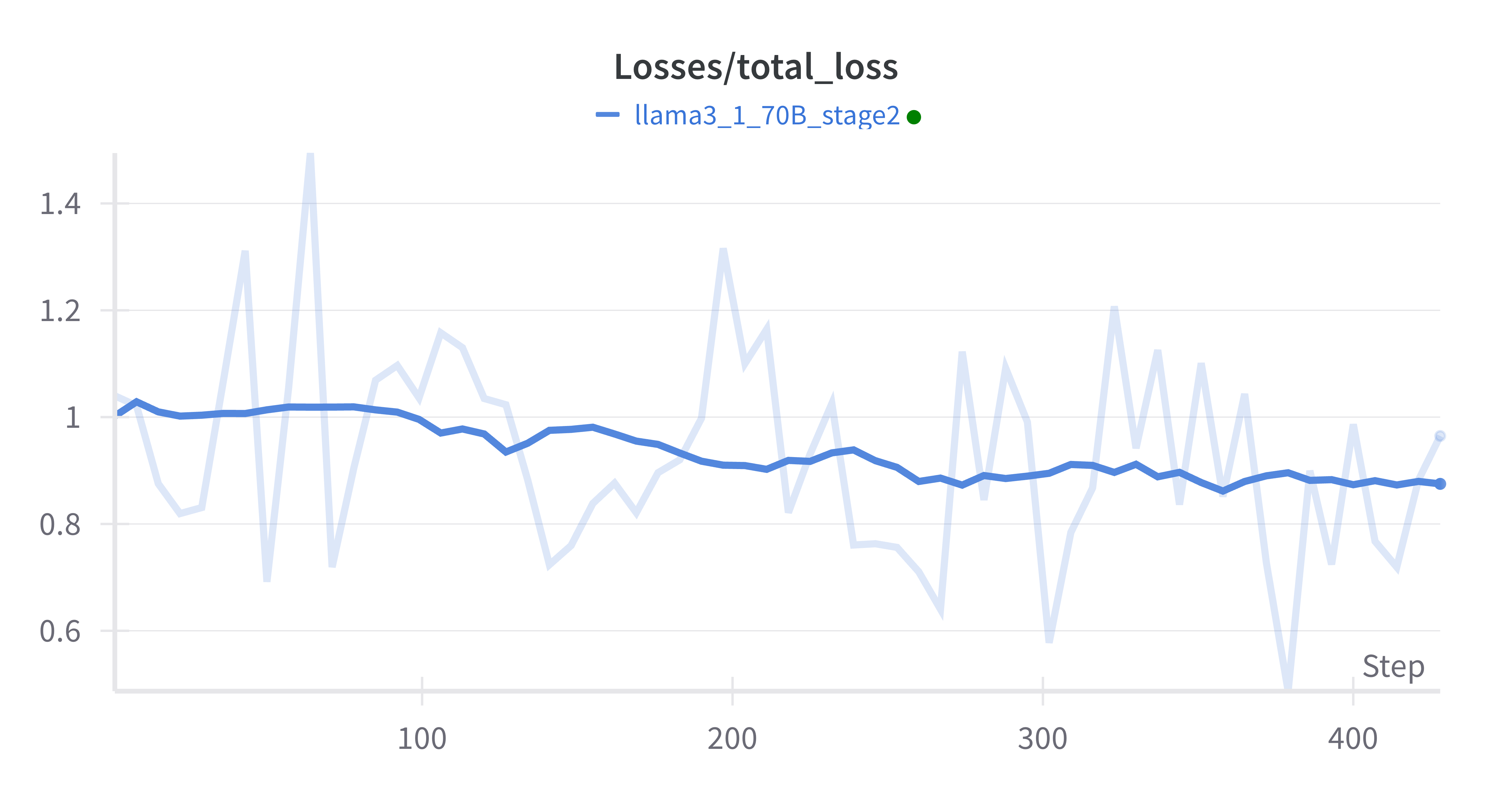}
        \caption*{(b) SFT Stage2 Loss}
    \end{minipage}

    \vspace{\baselineskip}

    \begin{minipage}{0.45\textwidth}
        \centering
        \includegraphics[width=\textwidth]{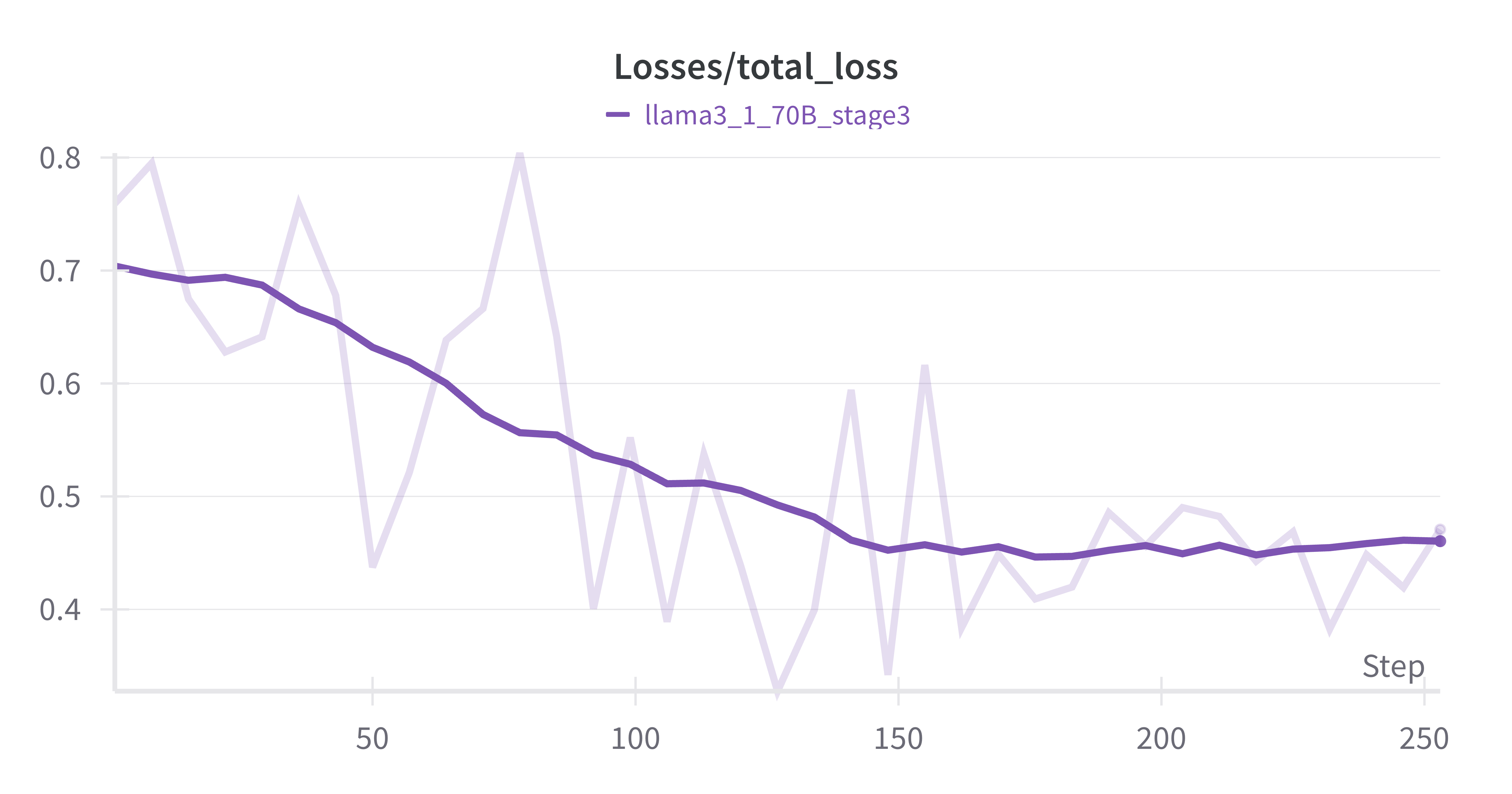}
        \caption*{(c) SFT Stage3 Loss}
    \end{minipage}
    \hfill
    \begin{minipage}{0.45\textwidth}
        \centering
        \includegraphics[width=\textwidth]{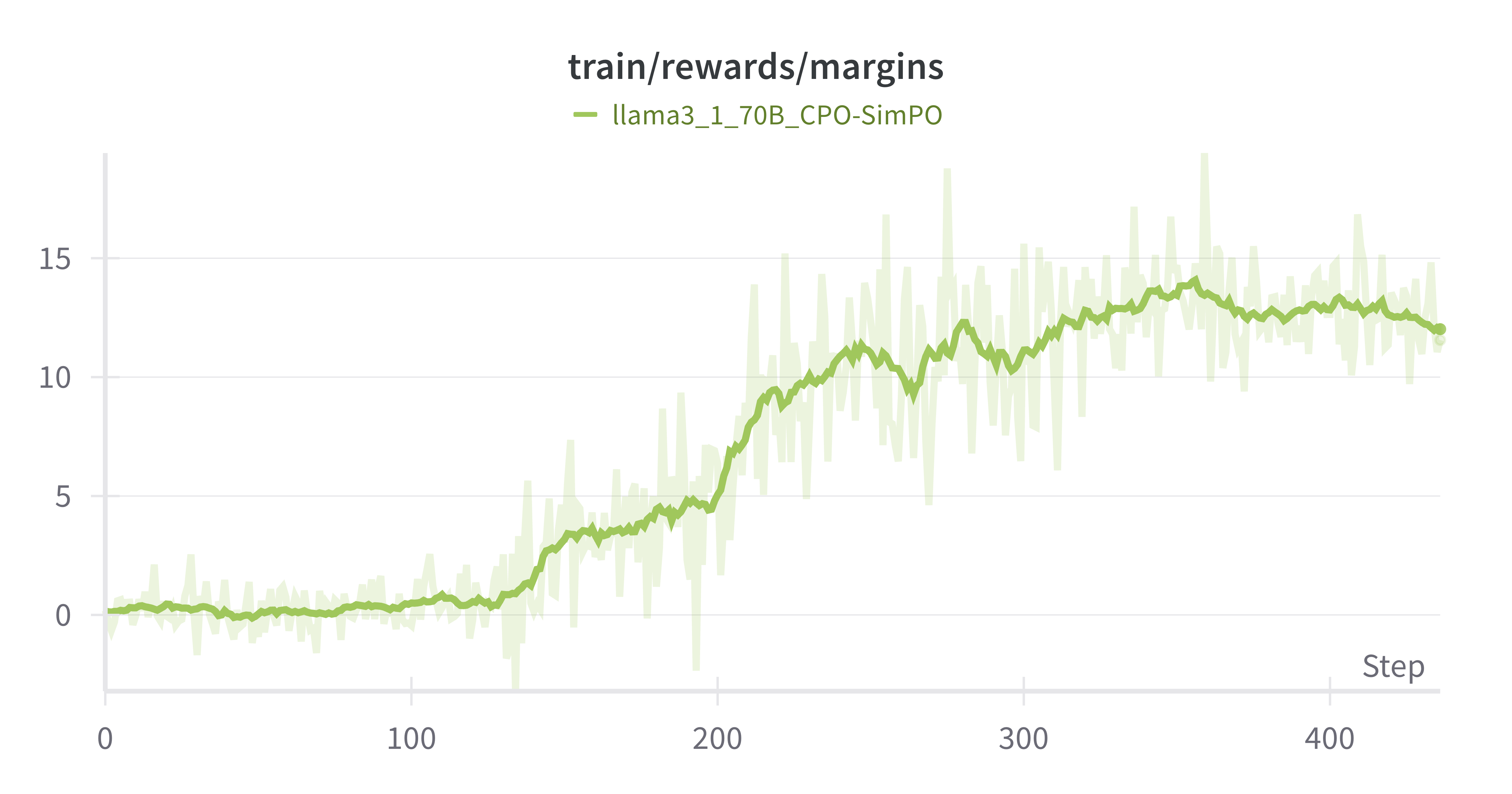}
        \caption*{(d) CPO-SimPO Rewards}
    \end{minipage}
    
    \caption{Training Loss. The figure shows the loss curve of each stage in the training process. The model gradually converges at each SFT stage in (a)(b)(c). In figure7(d), the alignment phase reward curve of CPO-SimPO gradually rises and converges.}
    \label{fig:Figure7}
\end{figure}

According to the Main Result Table.\ref{tab:Table4}, Citrus1.0-Llama-70B reach a top class performance on 70B scale models, especially on MedQA,PubMedQA,MedBullets,CareQA benchmark. Citrus also surpasses many close-source top LLMs such as Claude-sonnet and GPT-4o. Our model consistently demonstrates strong performance across a wide range of medical benchmarks, highlighting the effectiveness of our proposed approach.
Observing the loss curve in Figure.\ref{fig:Figure7}, it can be seen that the model gradually converges at each SFT stage. In the alignment phase, the reward curve of CPO-SimPO gradually rises and converges. The evaluation results indicate that the performance of the aligned model is the best among all stages.

\subsection{Future Discussion}
\begin{table*}[ht!] \centering 
\adjustbox{max width=1.0\linewidth}{
\begin{tabular}{lc} \toprule 
& \textbf{MedQA} \\
\midrule
\multicolumn{2}{c}{\bf \it Baseline LLMs} \\  
Llama-3.1-70B-base & 48.95  \\ 
Llama-3.1-70B-instruct & 78.40  \\
\midrule
\multicolumn{2}{c}{\bf \it Ours LLMs} \\
Citrus1.0-Qwen-72B & 87.12 \\
Citrus1.0-Llama-70B & 88.92 \\
\midrule
\multicolumn{2}{c}{\textit{{Supervised Finetune (SFT)}}} \\ 
Citrus-Llama-70B    stage1/2 & 77.06 \\
Citrus-Llama-70B    stage1/2 + stage3 w/ 20k sft data & 83.12 \\
Citrus-Llama-70B    stage1/2 + stage3 w/ 60k sft data & \underline{84.13} \\ 
Citrus-Llama-70B    stage1/2 + stage3 w/ 130k sft data & 83.74 \\    
\midrule 

\multicolumn{2}{c}{\textit{{Reinforcement Learning (RL)}}} \\ 
Citrus-Llama-70B-stage3   CPO+SIMPO  w/ 17k Rejection Sample data & 87.82 \\ 
Citrus-Llama-70B-stage3   CPO+SIMPO  w/ 45k Rejection Sample data & 87.20 \\ 
Citrus-Llama-70B-stage3   CPO+SIMPO  w/ 45k + 5k non-med Rejection Sample data & \textbf{88.92} \\ 

\bottomrule
\end{tabular}}
\caption{The ablation experiments on \textit{Citrus1.0-Llama-70B}. The SFT stage results and RL stage results are shown in sequence to show different contributions to final perform of model from different stages. We also implemented several experiments to reveal the impact from data size and data portion. "w/o" and "w/" denote "without" and "with". \textbf{Bold} highlights the best scores in each segment.Use \textbf{MedQA} benchmark to evaluate the influence on different training stages and data sizes.}
\label{tab:Table5}
\end{table*}

In the ablation experiments, we explore the impacts on different stage of training, including SFT stage 1,2,3 and RL stage. As the most distinguishable and influential benchmarks for medical scenario challenges, MedQA is carefully selected as "North star" during our training procedure from the base model all the way to the final one. The results shown in Table.\ref{tab:Table5} provide insights into the importance of each procedure in the training pipeline, discussed as below.

\paragraph{SFT Training Stages} The first two stages of SFT primarily focus on grounding the model with general knowledge and reasoning tasks. Training on these stages reach an acceptable level for the model to handling reasoning tasks. The third stage is where the model’s medical reasoning capabilities are fine-tuned.This stage's effectiveness is demonstrated in the performance improvements on the MedQA benchmark as the model progresses from 77.06 to 84.13.

\paragraph{SFT data size impact} The influence of data size on the model’s performance is evident when considering the results from different configurations of Stage 3. Fine-tuning with 20k SFT data yields a performance of 83.12, whereas utilizing 60k SFT data boosts performance to 84.13. However, further increasing the data size to 130k results in a slight performance dip to 83.74, suggesting diminishing returns as the model approaches an optimal configuration for this stage.

\paragraph{RL Data Proportion}
We experimented with varying the composition of the rejection sampling data. The most successful configuration involved using 45k medical questions and an additional 5k non-medical questions, resulting in a performance of 88.92. This configuration demonstrates that introducing non-medical question data in other scientific fields into the training process can help balance the model’s understanding of both domain-specific and general reasoning tasks, enhancing overall model performance.

\paragraph{RL Data Size Impact}
We also explore different RL data size on 17k and 45k. For 45k scale, we use the core, most-difficult 17k data combined with 28k other data, which is not challenging enough for complex medical cognitive task from same sources. The results show that the use of 17k rejection sample data yields a better performance 87.82, which is slightly higher than the model training by larger dataset improvements. This illustrated that a smaller size is enough for the model to capture the core ability of medical reasoning.

\section{Conclusion}
We present Citrus, a medical language model designed to enhance medical reasoning by emulating the cognitive processes of medical experts. Through a novel data synthesis approach and a multi-stage training methodology, we have developed a model capable of efficiently handling complex medical decision-making tasks. By releasing the model and its training data, we aim to promote further research in AI-driven medical reasoning and decision-making, thereby contributing to the advancement of healthcare technologies.

\paragraph{Thinking like an Expert}
We have constructed a medical reasoning dataset modeled from the cognitive processes of doctors, and have effectively demonstrated that such data significantly enhances the problem-solving capabilities of LLMs in medical scenario. Through an exploration of doctor\'s  thought processes, design of experimental data and attempts at model training, we have ultimately developed an LLM capable of effectively leveraging Long COT generated data to address medical issues. From a high-order perspective, we envision that our approach could be widely applicable across domains. By deconstructing the cognitive strategies of experts and utilizing representative core data to generate training data through our approach, models could potentially learn to abstract thinking specific to a given domain. We believe this approach can serve as a comprehensive alternative to human feedback. As a criterion, it effectively replaces the necessity of human feedback in training, allowing the model to understand the underlying characteristics of thinking. Through this understanding, the model can approach generalizable problems from an elevated level of cognitive abstraction, thereby becoming a domain expert.

\paragraph{Complex Training Pipeline}
We developed a comprehensive multi-phase training pipeline for Citrus, incorporating CPT, SFT, and RL, to enable the model to efficiently learn and adapt to complex medical reasoning tasks. By understanding the problem-solving thought processes of medical experts, we identified the dual-process theory and applied distinct cognitive strategies to various training phases using CPT and SFT. While we believe that extensive pre-training data and clinical examples will help the model perform pattern recognition, there is currently no effective method to equip the model with the complex reasoning abilities that medical experts use to solve problems. We employed a warm-up training phase using data courses and a carefully designed COT data generation method. By training the base model in a specific order, it is gradually enhanced into a medical reasoning model. In the final stage, we claim that offline RL training could further enhance the model's reasoning ability, ultimately ranking it among the top models of similar parameter scales on several authoritative benchmarks.

\newpage

{
\bibliographystyle{unsrt}
\bibliography{custom}
}


\clearpage
\appendix
\section{Ethical Statement}

Although the proposed model is a medical LLM with complex reasoning capabilities, it may still produce content that includes hallucinations or inaccuracies. Therefore, the current model is not suitable for real-world applications. Consequently, we will impose strict limitations on the use of our model. The models are not permitted for use in clinical or other industry applications where such inaccuracies could lead to unintended consequences. We emphasize the ethical responsibility of users to adhere to these restrictions in order to safeguard the safety and integrity of their applications.

\section{Prompt}
\label{ap-converquestion}

Here are the prompt examples.

\begin{prompt}[title={Reasoning Expert Prompt}]
{
You are tasked with addressing a medical examination question. Please carefully read the question, provide a detailed thought process, and then present your final answer.\newline

Here is the question:\newline
<Question>\newline
\textcolor{blue}{\texttt{\{Q\}}}\newline
</Question>\newline

Facing on the previous question, you are an assistant that engages in extremely thorough, self-questioning reasoning. \newline
Your approach mirrors human stream-of-consciousness thinking, characterized by continuous exploration, self-doubt, and iterative analysis.\newline
With the expectation that when facing this medical issue, you will be able to apply professional medical reasoning methods, such as differential diagnosis, to further reason and think about the problem.”\newline

Below is the definition of the differential diagnosis method in medical reasoning:\newline
Differential diagnosis refers to the process of systematically considering different possible diseases, ruling out diagnoses that do not match the condition, and ultimately determining the most likely disease. It involves the following steps:\newline
• Collecting information: Inquire about the medical history, conduct physical exams, and perform necessary laboratory tests.\newline
• Listing possible diagnoses: Based on the medical history, signs, symptoms, and laboratory results, list all possible diseases.\newline
• Gradually eliminating: Through further tests, symptom evaluations, and diagnostic tests, gradually eliminate impossible diagnoses, ultimately confirming the most likely disease.\newline
Differential diagnosis is a process of comparison and contrast, where doctors judge each potential diagnosis based on its characteristics, finding the disease that most closely matches the patient’s symptoms and signs.\newline

Please establish the following process in your logical reasoning:\newline
    1.	List all the known information in the problem, including the complete medical history and all test results.\newline
    2.	List possible diagnoses.\newline
    3.	Attempt to build a logical reasoning process.\newline

Below are the reasoning requirements; please ensure each step of the reasoning process meets the following criteria:\newline
(more details are listed in \url{https://github.com/jdh-algo/Citrus})\newline
            
Your reasoning steps should follow these requirements:     \newline     
<Reasoning>\newline
[Your extensive internal monologue goes here]\newline
- Begin with small, foundational observations\newline
- Question each step thoroughly\newline
- Show natural thought progression\newline
- Express doubts and uncertainties\newline
- Revise and backtrack if you need to\newline
- Continue until natural resolution\newline
</Reasoning>\newline
Please review the question again:\newline
<Question>\newline
\textcolor{blue}{\texttt{\{Q\}}}\newline
</Question>\newline
Please review the output format requirements again; your reasoning steps should be formatted as:\newline
<Reasoning>\newline
[Insert your reasoning step here.]\newline
</Reasoning>          \newline  
Please follow this output format for the next valid and useful reasoning step:\newline
}
\end{prompt}

\begin{prompt}[title={Reflection Expert Prompt}]
{
Please, as a very professional doctor, you will review the thought process of an ordinary doctor regarding a specific medical issue.\newline
You will see the question <Question>,  the previously established thought process<Previous Thought> and current reasoning step <Current Reasoning Step>\newline
Most importantly, you know the final answer <Ground Truth>. \newline
Please carefully evaluate, from an objective and professional perspective, whether the doctor’s reasoning step is logically sound.\newline
You may think carefully, step by step, and provide rigorous reasoning to argue whether this reasoning step is logically valid.\newline
Finally, you should rate its effectiveness with either 0 or 1, where 0 represents invalid, and 1 represents valid.\newline
Regardless of whether the reasoning is valid or not, please provide your feedback.\newline
If it is valid, please explain the logical reasoning form that is correct and suggest more possible directions for thinking.\newline
If it is invalid, please point out the flaws in the reasoning and provide a revised thought direction.\newline
The feedback should be heuristic, and you may guide them towards the correct answer.\newline
Below are the question and knowledge:\newline
<Question>\newline
\textcolor{blue}{\texttt{\{Q\}}}\newline
</Question>\newline
\textcolor{blue}{\texttt{\{GT\}}}\newline           
Below is the previously established thought process:\newline
<Previous Thought>\newline
\textcolor{blue}{\texttt{\{previous\string_thought\}}}\newline
</Previous Thought>\newline
Below is the current reasoning step:\newline
<Reasoning Step>\newline
\textcolor{blue}{\texttt{\{reasoning\string_step\}}}\newline
</Reasoning Step>\newline
The output format should strictly follow the format below:  \newline         
<Feedback>\newline
step feedback\newline
</Feedback>\newline
<Rating>\newline
1\newline
</Rating>\newline
Please consider whether the current reasoning step is positively helpful in answering the question, and remember to follow the output format at the end by providing feedback in a concise and precise text form, followed by the rating (0 or 1).\newline
You can include the Ground Truth in the feedback to provide necessary heuristic guidance, but do not mention terms like Ground Truth, Answer, etc., in the feedback. \newline       
Please use English for output:\newline
}
\end{prompt}

\section{Standard Inquiry Process}
\label{omc_pipeline:omc} 

\paragraph{User-submitted consultation requests} Users submit information regarding their medical conditions, symptoms, and medical history on the JDH platform. This information is typically submitted in the form of text, images, or videos.

\paragraph{Conversation records between doctors and users} After receiving a user’s consultation request, doctors engage in real-time text, voice, or video communication with the user. These conversation records contain key information such as the doctor’s inquiries about the user’s condition, the diagnostic process, and treatment recommendations.

\paragraph{Diagnostic plans and prescriptions} Based on the user’s condition description and conversation content, doctors provide diagnostic results, treatment plans, and medication prescriptions.

\section{Quality-Check}
\label{Quality-Check:qc} 
\paragraph{Consultation Professionalism}
\begin{itemize}
\item Assess whether the physician has omitted any critical or non-critical questions during the consultation regarding the patient’s chief complaint, current medical history, or past medical history, which could lead to insufficient grounds for the final conclusion.
\item Evaluate whether the physician failed to inquire about allergy history or special past medical conditions when recommending antibiotics or other treatments, potentially causing significant harm to the patient’s physical and mental well-being.
\item Determine whether the physician neglected to routinely collect information on the patient’s allergy history, liver and kidney function, or special disease history when advising on medications or products that may cause allergic reactions or other harm. Consultation quality must be comprehensive, detailed, and without any deficiencies.
\end{itemize}

\paragraph{Diagnostic Professionalism}
\begin{itemize}
\item Check for diagnostic errors by the physician.
\item Identify if only secondary diseases were diagnosed while primary diseases were overlooked.
\item Ascertain if the physician provided only a broad diagnosis when the patient’s description allowed for a more specific subtype diagnosis.
\item Ensure that the diagnostic basis is sufficient. Diagnostic quality must be comprehensive, accurate, and well-supported by evidence.
\end{itemize}

\paragraph{Question Professionalism}
\begin{itemize}
\item Verify the high relevance of the question content to the dialogue content.
\item Evaluate the rationality of the distractor options and whether they have a hierarchical relationship with the correct option, leading to ambiguity in the answer.
\item Ensure the information in the question is sufficient for diagnosis. The quality of the question must meet the standards of being informative, well-evidenced, and having reasonably set options.
\end{itemize}

\end{document}